%% file: Schomaker-Monk-field-test-2019-v0.1.tex
\documentclass[twoside,11pt]{article}
\usepackage{jmlr2e}

\usepackage{here}
\usepackage[usenames,dvipsnames,svgnames,table]{xcolor}
\usepackage{hyperref}

\jmlrheading{1}{2019}{x}{xx}{xxx}{Lambert Schomaker}

\ShortHeadings{A large-scale field test on word-image classification}{L. Schomaker}
\firstpageno{1}

\newlength{\figw}
\begin{document}

\title{A large-scale field test on word-image classification in large historical document collections
using a traditional and two deep-learning methods}

\author{\name Lambert Schomaker \email l.r.b.schomaker@rug.nl \\
       \addr Dept. of Artificial Intelligence  \\
       University of Groningen\\
       Nijenborgh 9, NL-9747 AG, The Netherlands}
       
\editor{~}

\maketitle

\begin{abstract}
This technical report describes a practical field test on word-image classification
in a very large collection of more than 300 diverse handwritten historical manuscripts, 
with 1.6 million unique labeled images and more than 11 million images used in testing.
Results indicate that several deep-learning tests completely failed (mean accuracy 83\%).
In the tests with more than 1000 output units (lexical words) in one-hot
encoding for classification, performance steeply drops to almost zero percent
accuracy, even with a modest size of the pre-final (i.e., penultimate) layer (150 units). 
A traditional feature method (BOVW) displays a consistent performance over numbers
of classes and numbers of training examples (mean accuracy 87\%). 
Additional tests using nearest mean on the output of the pre-final layer of an Inception V3 network, for each book,
only yielded mediocre results (mean accuracy 49\%), but was not sensitive to high numbers of 
classes. Notably, this experiment was only possible on the basis of
labels that were harvested on the basis of a traditional method which already works
starting from a single labeled image per class. It is expected that the performance
of the failed deep learning tests can be repaired, but only on the basis of human handcrafting (sic)
of network architecture and hyperparameters. When the failed problematic books are
not considered, end-to-end CNN training yields about 95\% accuracy. This average is
dominated by a large subset of Chinese characters, performances for other
script styles being lower. 
\end{abstract}


\section{Introduction}

This technical report\footnote{Version 0.1, 16th April 2019} 
describes the large-scale deployment of deep learning methods
in the Monk~\cite{Zant2008,ISR2009,vanOosten2014,DesignConsSchomaker2015},
project for lifelong machine learning in historical handwritten
collections. The promise of deep learning, as expressed implicitly in some
recent papers is that handcrafting is undesirable, 'traditional' 
and a thing of the past. Indeed, it would be extremely useful if a wide
range of problems can be handled using autonomously operating deep 
learning methods. Unfortunately, the great successes of today are the product
of another type of handcrafting, namely the tuning of hyperparameters, manual
design of network architectures as well as customized training procedures with
specifically designed loss functions, by human experts.

Therefore, the question we posed was: Is it really possible to have high performances
as are currently reported on academic and contrived benchmarks, but now on real 
data in large ongoing operations in the classification of (word) images in handwritten
historical collections? If the claim is true, human handcrafting should not
be necessary, given a decent deep learning architecture and training it
end to end. Instead of squeezing out fractional improvements on an often
used academic data set, this study is more in vain with the methods in
robotics, where systems are challenged under various operating conditions
(even being kicked by a technician) in order to study their robustness.

\subsection{Scale of the problem}

\begin{figure}[H] \center
\includegraphics[width=\figw]{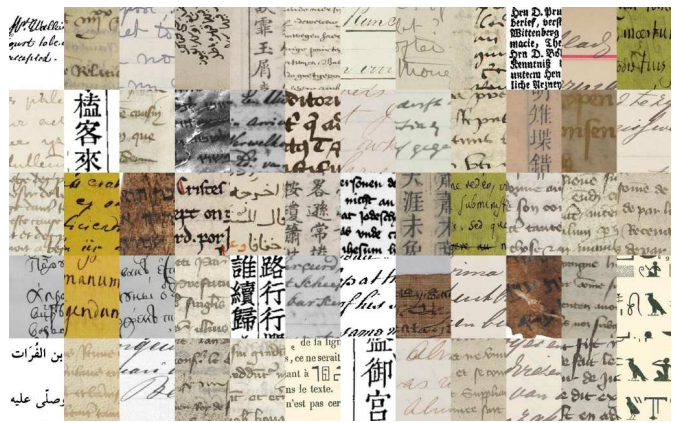} 
\caption{\label{fig:CollectionThumbsTop} Image samples from documents in the Monk collection}
\end{figure}

The trainable search engine for handwritten text in historical collections,
Monk~\cite{Zant2008,ISR2009,vanOosten2014,DesignConsSchomaker2015},
is an e-Science service for humanities researchers, archives and libraries that have
large numbers of scanned handwritten manuscripts, only indexed at the level of abstract
metadata and not in terms of textual content. As of April 2019, it contains 567
documents, 152477 pages, 147000 lexical words and shape classes, more than a 
million human-labeled training examples and about 700 million word-zone candidate
images. The handwriting styles vary from Western to Chinese, from Dead Sea Scrolls
and medieval to contemporary text. Additionally, a number of 'difficult' machine-printed 
documents are present, German fractur, Arabic printed text and Egyptian hieroglyphs.
The Chinese documents are mainly from the Harvard Yenching collection (305 manuscripts,
28900 pages). Figure~\ref{fig:CollectionThumbsTop} shows image samples from the Monk
collection. Word segmentation is assumed to be performed in this field test in earlier
processing stages.

\subsection{Research questions}

In view of the actual impressive successes~\cite{KrizhevskyImageNet2012,DeepMindGoRL2018} and the claims of success in the 
current deep-learning literature on the basis of long-existing academic data sets~\cite{MNIST1998,IAM2002,GW2012} 
or self-defined tasks~\cite{Gatys_2016_CVPR_GANpainter,DeepAtari2013}, 
a number of questions are relevant to existing operational large-scale systems, such as Monk.

\begin{enumerate}
\item What is the actual average performance of end-to-end trained
deep learning methods on a given set of highly diverse image-classification problems?

\item What is the actual average performance of the bottleneck vector of a pre-trained
deep learning method on such data?

\item How do these performances compare to a traditional bag-of-visual words (BOVW) feature method?

\item Is it true that human intervention (handcrafting) is not necessary anymore?

\end{enumerate}

\section{Method}

The goal is to perform a series of separate tests on a large set of books (documents)
in an automatic manner, i.e., by just starting a script that runs over the instances
in the set, uses a training set to train the classifiers, uses a validation set, and
ultimately the test set for that book, after which the next book is processed, etc.
Individual manual inspection and tuning was not allowed, per book.
Computing was done on two systems: A desktop system with a K620 NVIDIA GPU and an HPC
compute node with a V100 GPU. The experiment was started on January 4th 2019.

\subsection{Data}

On April 7th 2019, after 93 days of computing, 341 books were processed out of 567.
These 341 are the basis for the current evaluation.

From the existing labeled image set of each book, a sampling was made to
create a training set of 7/9th, a validation set of 1/9th and a test set of 1/9th
of the labeled data per book. Training set samples were augmented with five 
extra samples that were based on random elastic morphing~\cite{BulacuAugmentation2009,ImageMorphGithub}.
Only classes with a minimum of 20 training samples were used, only books with minimally
10 classes were used. Training sets are not mutually exclusive between books: Overlap
is possible. The reason is that in order to exploit style similarity, Monk uses a 
common training set for sibling books in the same style and language, while ensuring 
that the book-specific word and pattern classes are also part of the training set for
that book. Test instances and validation instances belong to the book, proper.
Table~\ref{tab:statClass} shows the statistics for the number of classes over the
books.

\begin{table}[H] \centering
\caption{\label{tab:statClass} Statistics of the number of classes over all books}
\medskip
\begin{tabular}{|l|l|} \hline
Classes & N \\ \hline
mean      &       1207.09 \\
min       &       10 \\
max       &       24421 \\
sd        &       2186.71 \\ \hline
\end{tabular}
\end{table}

\begin{figure}[H] \center
\includegraphics[width=\figw]{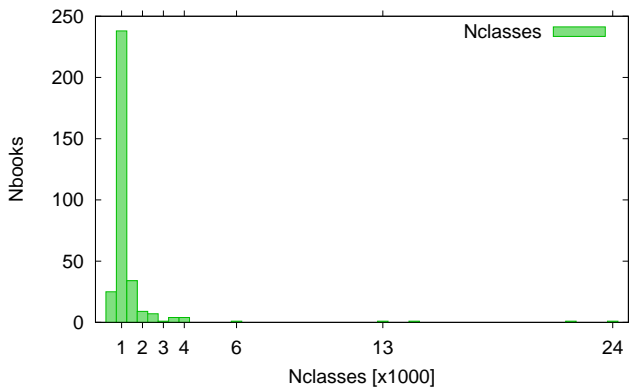} 
\caption{\label{fig:CNN-QP-histogram-Nclass}Histogram of number of books per number of classes}
\end{figure}

Figure~\ref{fig:CNN-QP-histogram-Nclass} shows the histogram of number of books per number of classes,
for all data sets (books). The modal number of classes is about 1000 lexical words or shapes.
Three exceptional cases have 13k,22k and 24k classes. The number of classes in each
book corresponds to the dimensionality of the one-hot encoded network output vector for classification.


The total (non-unique) number of human-labeled images over all training sets was 11 million. It should
be noted that training samples are assembled for a book on the basis of sibling books in the same style
and language.


The total number of {\em unique} human-labeled images in all training sets was 1.6 million.

In total 7 million images were added for data augmentation, using elastic random morphing~\cite{BulacuAugmentation2009} over
all books. In the assembling of training sets for each book, the total number such instances was 9 million due to overlap.



\begin{table}[H] \centering
\caption{\label{tab:nclass} Average number of classes per book in the major script groups}
\medskip
\input{nclass.tab}
\end{table}

\begin{table}[H] \centering
\caption{\label{tab:nhum} Average number of human-labeled instances per book in major script groups}
\medskip
\input{nhum.tab}
\end{table}

\begin{table}[H] \centering
\caption{\label{tab:nartif} Average number of randomly morphed augmented instances per book in major script groups}
\medskip
\input{nartif.tab}
\end{table}

\begin{table}[H] \centering
\caption{\label{tab:ntot} Average number of instances in the training set per book for major script groups}
\medskip
\input{ntot.tab}
\end{table}

Table~\ref{tab:nclass} shows the average number of classes per book in major script groups. The large group of Chinese documents (260)
will have the strongest influence on the results. These are mostly wood-block printed characters, but also handwritten and
metal font machine-printed documents.
Table~\ref{tab:nhum} shows the average number of human-labeled training instances per book in major script groups.
Note that the total number of unique labels (1.6 million) is less due to overlapped use between books.
Table~\ref{tab:nartif} shows the average number of augmented instances per book in the training set, in major script groups.
Random elastic morphing is used, generating believable (legible) word instances. No image flipping etc. was performed
as this is not suitable for handwriting and text patterns.
Table~\ref{tab:ntot} shows the average number of total instances per book in the training set for major script groups.
All data are stored on a single-mount point file system, from 2009-2018 this was GPFS, which was superseded by the clustered
file system Lustre in 2019~\cite{DesignConsSchomaker2015}.

\subsection{Classification method I: Bag of Visual Words, BOVW}

This is the baseline method within the Monk system. It was developed in 2008 and
consists of a computing a bag of visual words for several handwriting style groups.
It is computed by means of a Kohonen self-organized map. In the ingest phase of
unseen books, i.e., at the moment an archive or institution has uploaded a new book,
the best fitting BOVW is selected. In the current state of the system,
it is rarely necessary to compute a new BOVW. At the moment of performing this field
test there were 18 BOVW for the 567 books in use. No additional training was done.
The 'training' consists of computing the average BOVW for a (word) class.
Input word images are used, as is, after down scaling from 300dpi to 150dpi, or a 
comparable scale, whenever photographed images were used instead of flatbed scans.
Note that this method is also the bootstrapping method. It was used, together
with a method for providing intuitive hit lists by image matching~\cite{vanOosten2014},
to present hit lists to human users who flagged correctly recognized instances.
Not all samples were labeled in this manner, also impromptu labeling of pre-segmented
words in lines of text was used. It should be noted that the other two methods
indirectly benefit from this traditional approach, which can be applied immediately
upon the event of receiving a single image label.

\subsection{Classification method II: CNN, end-to-end}

The example Python program from the Tensorflow tutorial 'image classification' was
adapted to the current setting. The following architecture was used (Table~\ref{tab:CNNarch}):

\begin{table}[H] \centering
\caption{\label{tab:CNNarch} Deep network architecture, implemented in Tensorflow/Keras.}
\medskip
\begin{tabular}{l}
\verb+Input(wpix=100,50)-Lay1=32x(3x3)relu-Lay2=pool(3,2)+ \\
\verb+-Lay3=32x(3x3)relu-Lay4=pool(2,2)+ \\
\verb+-Lay5=24x(3x3)relu-Lay6=pool(1,1)+ \\
\verb+-Lay7=Flatten+ \\
\verb+-Lay8=FC(150)-relu-xNclasses=nnn-sigmoid+ \\
\verb+-opt=adam-loss=binary_crossentropy+ \\
\verb+-metrics=accuracy-nepochs=15+ \\
\end{tabular}
\end{table}

Input images were downscaled to 100x50 pixels. This scale will allow for
a 50-60\% word-classification accuracy in many collections using plain image template
matching (nearest neighbor, Pearson r). Therefore, it is to be expected that a 
deep-learning method should outperform this base rate drastically.
The architecture was based on separate experiments with handwritten
samples, without extensive experimentation, avoiding architecture handcrafting.
There were 15 epochs, which may seem low, but it needs to be kept in mind that
the test will be performed more than 300 times. This also precludes k-fold
evaluation. The gradient-descent methods used is Adam~\cite{KingmaAdamOpt2015}, with
a learning rate of 0.001, $beta_{1}=0.9, beta_{2}=0.999$. The number of 
units in the fully connected (FC) pre-final layer was 150. This number was 
chosen in view of expected problems with the number of classes in the
one-hot encoded output layer~\cite{ShengOpenSet2018}. A number of
1000 units in the pre-final layer and a typical number of classes of
thousand words would lead to 1 million weights for this FC layer alone.
With the current choice, only 150k weights are needed from the pre-final
layer to the classification layer.

\subsection{Classification method III: CNN, nearest-mean classification on bottleneck vector of Inception V3}

Here the pre-trained Inception V3 network was used and the output of the pre-final (penultimate)
layer was used to 'train' a nearest-mean classifier for each problem (book) at 
hand\footnote{A slightly adapted version of \url{https://www.tensorflow.org/hub/tutorials/image_retraining} was used.}.
The pre-final layer (FC with a one-hot class vector), has 2048 units. 
For some reason it is called the bottleneck layer, although this name is
usually used for the bottleneck hidden layer with the smallest number of units 
in an autoencoder and the dimensionality here (2048) is
rather high to represent a bottleneck. The term 'embedded' also does not appear
to be completely appropriate, since this learned representation does not represent the usual trained
low-dimensional subspace, as is the case in autoencoders and, e.g., the Word2vec~\cite{MikolovWord2Vec2013} algorithm. 
In related work we therefore
proposed the term 'tapped feature vector'~\cite{TappedFeatures2018}, because it 
is tapped from an existing system, as in electronics.
In the training, a mean vector (centroid) is computed for each class 
and nearest-centroid search is performed
for the classification, using Euclidean distance. The minibatch size is 100
images, the number of training steps 4000. For converting the images to tapped feature vectors,
the same training set was used as in the other two classification methods. The resulting
output file was split into a training part and a test part for nearest-mean matching. Two
partitionings were considered: odd/even (='50/50') and 7/8 train vs 1/8 test.

\subsection{What about LSTMs?}

In the subfield of handwriting recognition research that calls itself HTR (handwritten text recognition),
the goal is to transcribe a line-strip image of text. This is usually done with a long-short term
memory recurrent neural network (LSTM). In order to deal with geometric variations, often a deep CNN is required
as a front end, complicating the training. For linguistic post processing, common probabilistic tools are used that 
rely on text corpora of sufficient size. Although the LSTM is used in Monk, it is not applicable for this test. 
The different image problems (curvilinear lines, geometric variations) are too varied between books for this methodology.
Also, a linguistic reference corpus mostly does not exist yet: This is the whole purpose of the Monk system, to allow
newly scanned historical collections that are as yet without linguistic resources an entry into the digital
phase of their life cycle: A chicken-and-egg problem. In brief, the complications in training diverse individual
documents are - at the moment - too severe to apply LSTMs on unmonitored processing. Additionally, the current
goal is concerned with isolated word images, which is not the ballpark where LSTMs are expected to excel, due
to the rather limited temporal context. We leave this
as an exercise in the near future~\cite{GideonNoPadding2019}.

\section{Results}

\subsection{Computation times}

Table~\ref{tab:timing} shows the average computation time per book for different
methods and different computing hardware used (2 GPU types, 1 CPU). The use of
a V100 GPU is clearly desirable. Computations for the nearest-mean methods (CNN bottleneck and
BOVW) are directly determined by the number of classes.


\begin{table}[H] \centering
\caption{\label{tab:timing} Average computation time per book for different
methods and different computing hardware.}
\begin{tabular}{|l|r|l|r|}\hline
Hardware & Hours/book & Method & Nbooks \\ \hline
K620 &   8.5 & CNN end-to-end & 263 \\
V100 &   1.8 & CNN end-to-end & 272 \\
K620 &   2.8 & CNN bottleneck & 81 \\
CPU  &   4.1 & BOVW           & 208 \\ \hline
\end{tabular}
\end{table}


\subsection{Raw plots}

\begin{figure}[H] \center
\includegraphics[width=\figw]{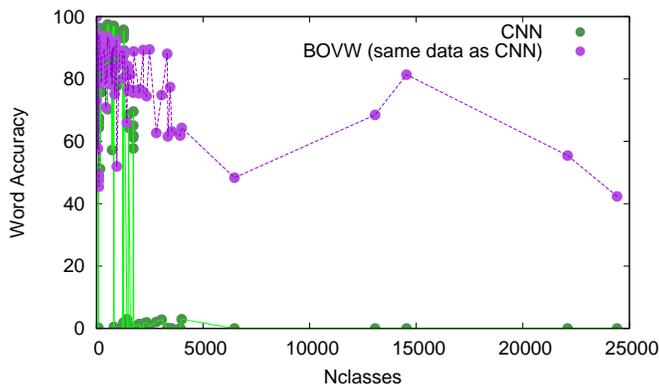} 
\caption{\label{fig:d-Perf-CNN-and-QP-per-Nclass} Word accuracy as a function of number 
of classes in a book. Each point corresponds to a document training and testing event. 
A green points represents a CNN test, a purple point represents a BOVW test.}
\end{figure}

Since this is a technical report, results will be first presented as raw plots
of dots with connected lines, the x-axis being sorted on increasing values.
A scatter plot might be considered more appropriate, but the lines make the
surprising aspects of the results more clear.
Figure~\ref{fig:d-Perf-CNN-and-QP-per-Nclass} shows word accuracy as a function of number of
classes in a book. A green points represents a CNN test, a purple point represents a BOVW test.
Deep learning (end-to-end trained CNNs) outperform BOVW below 1000 classes in a data set (book).
Above 1000 classes in a data set, the CNN-accuracy collapses to far below 10\% while the
traditional BOVW deteriorates in a less dramatic manner. However, also here a nearest centroid
matching only attains slightly over 40\% accuracy in case of 22k classes.

\begin{figure}[H] \center
\includegraphics[width=\figw]{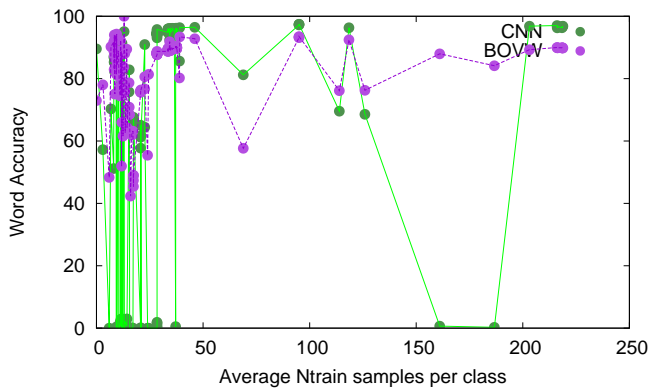} 
\caption{\label{fig:d-Perf-CNN-and-QP-per-AvgNtrain} Word accuracy as a function of number of average number of 
training samples per class in a book. A green points represents a CNN test, a purple point represents a BOVW test.}
\end{figure}

Figure~\ref{fig:d-Perf-CNN-and-QP-per-AvgNtrain} shows word accuracy as a function of number of average number of
training samples per class in a book. Below an average of 30 examples per class, several CNN tests clearly fail. 
In other cases the performance will
be higher for CNN than for BOVW. However, even at more than 175 examples per class, a CNN training may fail,
presumably because the number of output units, i.e., the number of image classes, is too high.

\begin{figure}[H] \center
\includegraphics[width=\figw]{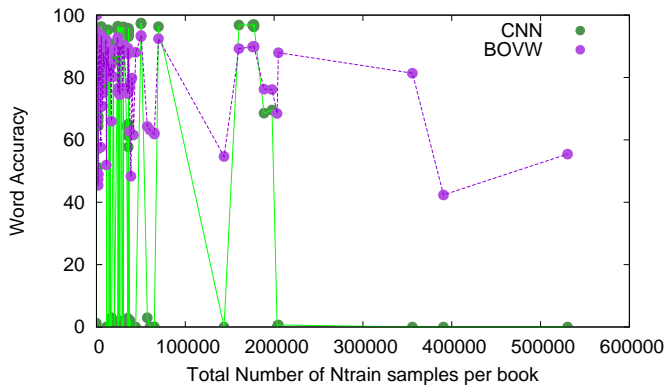} 
\caption{\label{fig:d-Perf-CNN-and-QP-per-Ntrain} Word accuracy as a function of number of training samples per book.
Color coding is as before.} 
\end{figure}

Figure~\ref{fig:d-Perf-CNN-and-QP-per-Ntrain} shows word accuracy as a function of number of training samples per book.
Apart from the 'Nclasses' problem, i.e., the risk of a failed training when the dimensionality of the one-hot encoded
output vector exceeds 1000 (classes), the CNN can display superior performances in many books. However, even 200000 
training samples did not help a CNN in case of a number of output classes that is too high. The BOVW method 
was able to profit from large numbers of training examples, yielding a performance above 55\% accuracy at 500k training 
instances.

\begin{figure}[H] \center
\includegraphics[width=\figw]{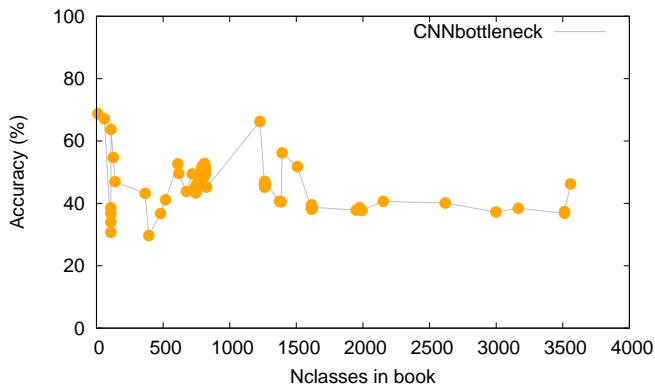} 
\caption{\label{fig:make-CNN-bottleneck} Word accuracy over number of classes for the 'bottleneck' feature vector
(Ndim=2048) of a pre-trained Inception V3 network.}
\end{figure}

Figure~\ref{fig:make-CNN-bottleneck} shows the word accuracy over number of classes for the 'bottleneck' feature vector
(Ndim=2048) of a pre-trained Inception V3 network, using nearest-mean classification. 
For each class in the training set, the mean vector was computed and
the test set was evaluated using the Euclidean distance between an unknown sample and the centroids in the training set.
Modal performance is about 50\%, which is evidently far above chance level given the large number
of classes in a book but much lower than the other two methods. As with the other nearest-mean method (BOVW),
the performance is less sensitive to the number of classes than the regular deep CNN, i.e., not collapsing after
about 1000 classes. For the odd/even evaluation, average accuracy is 48.9 ($\pm \sigma = 7.2$) \% with a mode of 50\%. 
For the 1/8 test vs 7/8 train evaluation, average accuracy is 48.3 ($\pm \sigma = 5.3$) \%, also with a mode of 50\%.


\vfill

\subsection{Performance histogram}

\begin{figure}[H] \center
\includegraphics[width=\figw]{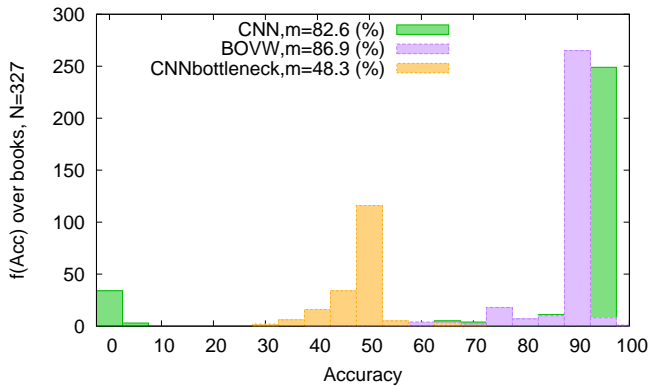} 
\caption{\label{fig:CNN-QP-histograms}Histogram of number of books per obtained accuracy
band, for end-to-end trained CNN (green), for a pre-trained CNN-based nearest mean classifier 
and for a BOVW-based feature (purple).}
\end{figure}

\begin{figure}[H] \center
\includegraphics[width=\figw]{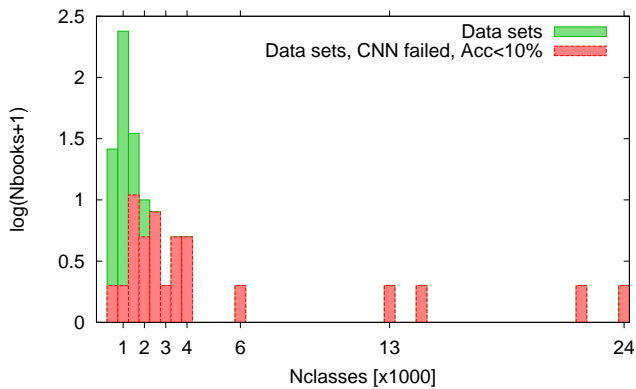} 
\caption{\label{fig:CNN-QP-histogram-Nclass-LOG}Histogram of log number of books per number of classes, 
for all data sets (green) and for CNN-based data sets with an accuracy below 10\% (red).}
\end{figure}

Figure~\ref{fig:CNN-QP-histograms} shows the distributions of books per obtained accuracy
band, for end-to-end trained CNN, for the pre-trained CNN-based 'bottleneck vector' with
nearest-mean classification and for a BOVW-based feature.
Although the deep-learning approach has a clear peak at approx. 95\% accuracy, whereas the
BOVW peaks at a lower 90\% accuracy, the average performance for the CNN is pulled down due to
a non-negligible number of books with a fully failed training. The reason for this is most likely
located in the number of output dimensions of the one-hot class vector. Figure~\ref{fig:CNN-QP-histogram-Nclass-LOG}
shows the histogram of $log(number of books+1)$ against number of classes. The figure is a variant of
Figure~\ref{fig:CNN-QP-histogram-Nclass}, in log scale, marking the failed CNN cases.
In order to avoid -infinity, a dummy count of 1 was added to be able to zoom in on the problematic
cases. With one exception the failed books have more than 1000 (word) classes. Table~\ref{tab:final} shows
the average accuracy for the three methods.


\begin{table}[H] \centering
\caption{\label{tab:final} Average accuracies in \%}
\medskip
\begin{tabular}{|l|r|r|} \hline
Method & m & sd \\ \hline
CNN           & 82.6 & 30.1 \\
BOVW          & 86.9 & 8.3 \\
CNNbottleneck & 48.4 & 5.2 \\ \hline
\end{tabular}
\end{table}

The large standard deviation of the accuracy in case of the end-to-end trained CNN
reconfirms the risk involved in unmonitored training of a deep-learning method 
in ongoing 24/7 operations.

\subsection{Second attempt of training}

We performed 51 cases of a second attempt of training the end-to-end CNN, of which
29 did not succeed (average number of classes 3072), 8 improved, i.e., yielded an accuracy above 10\% (average
number of classes 1429) and 14 books worsened from a good performance to a performance to below 10\% (average
number of classes 1292).




\section{Conclusion}

The purpose of this field test was not to present yet another fractional improvement
on an overused, contaminated academic benchmark. Secondly, the purpose was not to
question the success of deep-learning methods, per se. What is more important is 
to consider the macroscopic perspective where the number of data sets is huge and
their variation is very wide. Furthermore, it is more than likely that other variants 
of each of the three methods used in this field test would yield a higher accuracy performance 
if more human effort is spent on hyperparameters, architectures, features and attribute representations. 
Rather, we hope to have demonstrated that at a scale of more than three hundred image collections 
with a total of 9 millions images considered, it is not humanly possible to fine tune on individual
problems in, e.g., Western-medieval handwritten vs Chinese wood-block printed styles.
Each document poses a whole range of new problems that must be solved by computational
intelligence, not by humans.

Deep learning is still a product with the label 'assembly required'. Thinking
about optimal network architectures, hyperparameters and attribute representations
by human designers is still essential. Therefore, deep-learning methods require more human
overseeing than the traditional lean-training methods when they are embedded in a 24/7 learning
engine such as the Monk system, that operates largely autonomously since 2009. 
Even at a medium size of 150 units in the penultimate layer of an eight-layer CNN, the
probability of a failed training increases drastically beyond 1000 output units (classes)
in one-hot encoding. At the same time, nearest-mean approaches provide a stable 'training',
up to 22classes. The use of a pre-trained feature (the penultimate layer of Inception V3) proved
to be disappointing.

On the other hand, disregarding its failures in this field test, deep learning is attractive because of the
usually higher performance for the trainings that succeeded. Also, computational load
is attractive if an NVIDIA V100 is used for end-to-end training. It is a pity that
not all tests succeed. Retrying helps in some cases with Nclasses roughly between 1000 and 2000,
but on a retry, there is also a probability of failure. 

The challenge is to have the best of both worlds, making use of the 'ballpark principle'\footnote{This will be presented in a Chapter in the HISDOC book, in preparation}:
start with lean methods when the number of labels is low, harvest more labels on this basis,
in order to apply deep learning when enough labels are collected. For large-lexicon problems,
convenient script-specific attribute representations, such as PHOC~\cite{PHOCAlmazan2014,PHOCSudholt2016} 
or WUBI~\cite{ShengOpenSet2018} need to be adapted to the
language at hand. It may be hoped that after such a fundamental choice, deep learning can
indeed run autonomously without human inspection and intervention. Finally, the results
of this field test show how misleading the mere usage of average performances is.
As noted earlier~\cite{PenComp1998} (cf. Fig. 9), the use of performance histograms is
more informative and more realistic to get an impression of real-life operations than
a single mean value, even if standard deviations are mentioned. In the current study,
the books that failed on CNN appear on the left of the performance histogram (Figure~\ref{fig:CNN-QP-histograms}) and
should be left out, in order to see the real potential of this particular deep learning 
method. 

It is clear that new concepts for autonomous machine learning are needed. The term AutoML was coined by Guyon 
et al.~\cite{automlchallenges} in a series of challenge benchmarks for problems that need to be solved
without human intervention. It is clear that lifelong autonomous machine learning
is a necessity when both scalability and continuous variation are an issue, as in Monk, 
where there is a continuous stream of incoming books, where the image labels are in a continuous 
flux, both in terms of their total number and in terms of label substitution due to changing insights
of scholarly users.



\acks{We would like to acknowledge support for this project
from the Dutch NWO (grants 612.066.514 and 640.002.402) and SNN/EU project 'Target'}


\vskip 0.2in
\bibliography{Schomaker-Monk-field-test-2019-v0.1}

\end{document}

%% file: nclass.tab
\begin{tabular}{|l|r|r|r|} \hline
Group & mean & sd & Nbooks \\ \hline
17-19th-century & 3261.0 & 4244.5 & 24 \\
20th-century & 2194.2 & 3356.8 & 17 \\
Chinese & 1023.9 & 1975.2 & 260 \\
Medieval & 1415.0 & 938.3 & 27 \\
Misc & 666.4 & 807.9 & 17 \\ \hline
\end{tabular}

%% file: nhum.tab
\begin{tabular}{|l|r|r|r|} \hline
Group & mean & sd & Nbooks \\ \hline
17-19th-century & 17703.7 & 18932.6 & 24 \\
20th-century & 21581.7 & 55118.9 & 17 \\
Chinese & 29162.2 & 27889.0 & 260 \\
Medieval & 14968.1 & 11709.8 & 27 \\
Misc & 16003.8 & 17693.3 & 17 \\ \hline
\end{tabular}

%% file: nartif.tab
\begin{tabular}{|l|r|r|r|} \hline
Group & mean & sd & Nbooks \\ \hline
17-19th-century & 99337.8 & 328848.0 & 24 \\
20th-century & 15512.2 & 28359.1 & 17 \\
Chinese & 21461.0 & 153990.0 & 260 \\
Medieval & 25358.7 & 50874.9 & 27 \\
Misc & 19307.8 & 51669.3 & 17 \\ \hline
\end{tabular}

%% file: ntot.tab
\begin{tabular}{|l|r|r|r|} \hline
Group & mean & sd & Nbooks \\ \hline
17-19th-century & 117041.0 & 338817.0 & 24 \\
20th-century & 37093.9 & 83167.4 & 17 \\
Chinese & 50623.2 & 180432.0 & 260 \\
Medieval & 40326.8 & 57006.4 & 27 \\
Misc & 35311.5 & 55846.8 & 17 \\ \hline
\end{tabular}